\documentclass{Interspeech}

\interspeechcameraready

\title{CNVSRC 2024: The Second Chinese Continuous Visual Speech Recognition Challenge\thanks{Z. Liu and X. Li are joint first authors.}}

\author[affiliation={1}]{Zehua}{Liu}
\author[affiliation={1}]{Xiaolou}{Li}
\author[affiliation={2}]{Chen}{Chen}
\author[affiliation={1}]{Lantian}{Li}
\author[affiliation={2}]{Dong}{Wang}

\affiliation{School of Artificial Intelligence}{Beijing University of Posts and Telecommunications}{China}
\affiliation{Center for Speech and Language Technologies}{Tsinghua University}{China}

\email{\{lzh211,lixiaolou,lilt\}@bupt.edu.cn,\{chenc21,wangdong99\}@mails.tsinghua.edu.cn}


\usepackage{comment}
\begin{document}

\maketitle

\begin{abstract}
This paper presents the second Chinese Continuous Visual Speech Recognition Challenge (CNVSRC 2024), which builds on CNVSRC 2023 to advance research in Chinese Large Vocabulary Continuous Visual Speech Recognition (LVC-VSR). 
The challenge evaluates two test scenarios: reading in recording studios and Internet speech. 
CNVSRC 2024 uses the same datasets as its predecessor CNVSRC 2023, which involves \emph{CN-CVS} for training and
\emph{CNVSRC-Single/Multi} for development and evaluation. However, CNVSRC 2024 introduced two key improvements: 
(1) a stronger baseline system,  and (2) an additional dataset, \emph{CN-CVS2-P1}, for open tracks to improve data volume and diversity. 
The new challenge has demonstrated several important innovations in data preprocessing, feature extraction, model design, and training strategies, further pushing the state-of-the-art in Chinese LVC-VSR. 
More details and resources are available at the official website\footnote{https://cnceleb.org/competition}.

\end{abstract}

\section{Introduction}

Visual Speech Recognition (VSR), also known as lip reading, is a technology that deciphers speech content from lip movements. 
It has a wide range of applications, including public surveillance, assistive technologies for the elderly and disabled, 
and deepfake detection~\cite{goldschen1997continuous,tye2007audiovisual,hassanat2011visual,haliassos2021lips}. 
Despite significant advancements in word- and 
phrase-level recognition~\cite{yeo2023multi,ma2022training,yang2022improved,chung2017lip,kim2022distinguishing}, 
VSR research remains in its early stages and still faces challenges in real-world applications, particularly in the domain of large vocabulary continuous visual speech recognition (LVC-VSR)~\cite{chang2024conformer,serdyuk2022transformer,shilearning,shillingford2018large}.

This challenge is especially pronounced for the Chinese language, mainly primarily due to the scarcity of publicly available data resources. To address this, the CN-CVS dataset~\cite{chen2023cn} was released in 2023 as the first large-scale multi-modal dataset for Chinese visual speech recognition. This dataset has provided a strong foundation for advancing LVC-VSR research. 
Building on this progress, the first Chinese Continuous Visual Speech Recognition Challenge (CNVSRC 2023) reported several highly effective methodologies, including Chinese character-based modeling, extensive data augmentation, fully 3D-CNN visual front-ends, cross-modal modeling, and system fusion~\cite{chen2024cn}, further promoting research in this direction.


Building on the success of CNVSRC 2023 and its associated workshop, 
the second Chinese Continuous Visual Speech Recognition Challenge (CNVSRC 2024) was launched and 
hosted a workshop at NCMMSC 2024 for technical summary and discussion\footnote{https://cnceleb.org/workshop}.

Like its predecessor, CNVSRC 2024 aims to evaluate Chinese LVC-VSR systems in two primary test scenarios: reading in a recording studio and speech on the Internet. To ensure consistency in benchmarking and track technical progress, it fully followed the test protocol of CNVSRC 2023, maintaining the same configurations for the data profile and evaluation metric.

Meanwhile, CNVSRC 2024 introduces two key improvements based on the experiences and insights gained from CNVSRC 2023: (1) a more powerful baseline system was provided, reflecting the state-of-the-art performance on the task, and (2) an additional dataset, CN-CVS2-P1, was released for the open tracks. To facilitate research and development, all datasets, baseline models, and code have been open-sourced. This will enable participants to build their own systems easily and effectively. This paper provides a comprehensive summary of CNVSRC 2024, highlighting the representative techniques and insights gained from submitted systems.

The structure of the rest of this paper is outlined as follows: 
Section~\ref{sec:td} introduces the task settings and data profile. 
Section~\ref{sec:base} provides a comprehensive description of the baseline system, 
including the model structure, training strategies, and performance evaluation. 
Section~\ref{sec:res} reports the challenge results and summarizes the representative technologies utilized by the participants. 
Finally, Section~\ref{sec:conc} concludes the paper.

\section{Tasks and Data}
\label{sec:td}
The tasks in CNVSRC 2024 are the same as those in CNVSRC 2023~\cite{chen2024cn}, mainly divided into two categories: 1) tasks for a specific speaker scenario, and 2) tasks for non-specific speakers. Each scenario is further divided into two tracks: open track and fixed track. 
To further support model development for the open track, CNVSRC 2024 introduces CN-CVS2-P1, expanding data resources compared to CNVSRC 2023.
The specific details are as follows.

\subsection{Tasks}

CNVSRC 2024 consists of two tasks: Single-speaker VSR (T1) and Multi-speaker VSR (T2). 
The former (T1) emphasizes the performance of large-scale tuning for a specific speaker,
while the latter (T2) focuses on the basic performance of the system for non-specific but \emph{registered} speakers, i.e., speakers seen in the data for system development.
In both tasks, the system is fed with silent facial videos that contain a single person, 
and the system is required to generate the spoken content in the form of word sequences. Considering the present technical status, CNVSRC 2024 did not set up a fully speaker-independent track in this challenge.  

Furthermore, each task is categorized into a `fixed track' and an `open track'. 
For the fixed track, only the data provided by the organizer can be used, and the use of additional resources should be agreed upon by the organization committee. 
For the open track, any data sources and tools can be used to construct their systems, except for the data in the evaluation set. 

\begin{table}[htb]
\centering
\caption{Task description of CNVSRC 2024.}
\label{tab:task}
\resizebox{1\columnwidth}{!}{
\begin{tabular}{l|c|c}
  \toprule
                           & Fixed Track                   &   Open Track         \\
  \midrule
  ~~T1: Single-speaker VSR~~ & ~~CN-CVS, CNVSRC-Single.Dev~~ & ~~No constraint (e.g. CN-CVS2-P1)~~ \\
  \midrule
  ~~T2: Multi-speaker VSR~~ & ~~CN-CVS, CNVSRC-Multi.Dev~~ & ~~No constraint (e.g. CN-CVS2-P1)~~ \\
  \bottomrule
\end{tabular}}
\end{table}

CNVSRC 2024 uses Character Error Rate (CER) as the main metric to evaluate the performance, formulated as follows:

\begin{equation}
\label{eq:cer}
    CER = \frac{S + D + I}{N}
\end{equation}

\noindent where $S$, $D$, and $I$ represent the number of substitutions, deletions, and insertions in the output transcription, respectively. 
$N$ is the number of characters in the ground truth transcription.

\subsection{Data Profile}

CNVSRC 2024 employs the CN-CVS dataset as the training set, along with two additional datasets: 
CNVSRC-Single and CNVSRC-Multi, which serve as the development and evaluation sets for the two tasks: 
Single-speaker VSR (T1) and Multi-speaker VSR (T2) respectively. 
Additionally, CNVSRC 2024 provides an additional data source, CN-CVS2-P1, to facilitate enhancing the performance of \emph{open track} models. 
Table~\ref{tab:data} presents the data profile of these datasets. 
For both CNVSRC-Single and CNVSRC-Multi, the entire dataset is divided into a development set and an evaluation set. 
The development data is fully accessible to the participants (including video, audio, and text), 
whereas the text and audio of the evaluation data remain confidential during the entire challenge process. 
Participants can use the development data in any way, e.g., freely splitting it into a subset for model fine-tuning and a subset for model validation/selection.

\begin{table}[htb]
\centering
\caption{Data profile used in CNVSRC 2024.}
\label{tab:data}
\resizebox{1\columnwidth}{!}{
 \begin{tabular}{l|c|c|c|c|c|c}
 \toprule
 & \multicolumn{1}{c|}{CN-CVS} & \multicolumn{1}{c|}{~~CN-CVS2-P1~~} & \multicolumn{2}{c|}{CNVSRC-Single}  & \multicolumn{2}{c}{CNVSRC-Multi}  \\
 \midrule
 DataSet     & Train            &    Train         & Dev  & Eval              & Dev    & Eval   \\
 \midrule
 \# Videos~~   & ~~206,261~~      &  ~~160,116~~     & ~~25,947~~ & ~~2,881~~   & ~~20,450~~    & ~~10,269~~  \\ 
 \# Hours~~    & ~~308.00~~       &  ~~196.77~~      & ~~94.00~~  & ~~8.41~~    & ~~29.24~~     & ~~14.49~~   \\ 
 \bottomrule
\end{tabular}}
\vspace{-2mm}
\end{table}

\noindent \textbf{CN-CVS}: The CN-CVS dataset comprises visual-speech data from over 2,557 speakers, which amounts to more than 300 hours of data in total. 
It encompasses various scenarios, including news broadcasts and public speeches, 
making it the largest open-source Chinese visual-speech dataset available at present. 

\noindent \textbf{CNVSRC-Single}
The CNVSRC-Single dataset was designed specifically for Single-speaker VSR (T1) and was downloaded from a broadcaster's online channel. 
It includes over 800 video clips of that broadcaster, and the cumulative duration is over 100 hours. 
Nine-tenths of the data were released as the development set, while the remaining one-tenth was the evaluation set.

\noindent \textbf{CNVSRC-Multi}
The CNVSRC-Multi dataset was designed as the development/evaluation data for Multi-speaker VSR (T2). 
It involves two scenarios: reading in a recording studio and speech on the Internet. 
It includes audio and video data from 43 speakers, with nearly 1 hour of data per person. 
Two-thirds of each person's data were released as the development set, while the remaining data served as the evaluation set. 
The data from 23 speakers were recorded in a recording studio with fixed camera positions and a reading style, 
with each recording being relatively short. 
The data from the other 20 speakers were obtained from the internet, featuring longer recording durations and more complex environments and content.

\noindent \textbf{CN-CVS2-P1}
The CN-CVS2-P1 dataset serves as an additional data source for the open tracks. 
This data source is the first part of a newly designed dataset, named CN-CVS2, that was collected from the Internet and will be published shortly. 
CN-CVS2-P1 includes over 160,000 utterances, about 200 hours in total. Note that it is only permitted to be used in the open tracks of the challenge, not the fixed tracks.

\section{Baseline System}
\label{sec:base}

We developed two baseline systems based on the Auto-AVSR framework~\cite{ma2023auto}: 
one for the Single-speaker VSR task (T1) and the other for the Multi-speaker VSR task (T2). 
Only the datasets provided by the organizer (except CN-CVS2-P1) were used, meaning these systems conform to the specifications of the fixed tracks.

Building on the technical insights from CNVSRC 2023~\cite{chen2024cn}, the baselines of CNVSRC 2024 make two main improvements compared to CNVSRC 2023: 
(1) Replace Chinese subword units with Chinese character units;
(2) Implement the Attention decoder of the hybrid CTC/Attention architecture using a Bi-Transformer decoder instead of the standard forward Transformer. 
These improvements have yielded a substantial performance boost. More details are presented below.

\begin{table*}[htb]
\centering
\vspace{-5mm}
\caption{Training Details of the Pretraining (P1 \& P2) and Fine-tuning (FT) steps when constructing the baseline systems.}
\vspace{-2mm}
\footnotesize
\label{tab:training}
\resizebox{0.77\linewidth}{!}{ 
\begin{tabular}{l|c|c|c|c}
    \toprule
    Experiment        & P1                     & P2              & FT (Single-Speaker)              & FT (Multi-Speaker)                  \\
    \midrule                                                                                                     
    Initialize        & Random                 & P1 Saved Model  & P2 Saved Model  & P2 Saved Model      \\
    Warmup Epochs     & 5 & 5 & 2 & 2\\
    Learning Rate     & 0.0002                 & 0.001           & 0.0003          & 0.0002              \\
    Training Epochs   & 75 $+$ Early stop                     & 75              & 80              & 80                  \\
    Saved Model       & Top 10 average     & Last 10 epochs average & Last 5 epochs average  & Last 5 epochs  average     \\
    \bottomrule
\end{tabular}}
\vspace{-2mm}
\end{table*}

\subsection{Model Structure}
The model structure is adopted from Auto-AVSR~\cite{ma2023auto}. 
Specifically, it comprises three components: visual frontend, encoder, and decoder. 
Similar to the baselines of CNVSRC 2023, the visual frontend uses ResNet18~\cite{stafylakis2017combining} as its backbone, with one 3D-CNN layer followed by eight 2D-CNN layers to capture local spatiotemporal correlations. 
The encoder employs a Conformer structure~\cite{gulati2020conformer}, which involves 12 layers of Conformer blocks to extract context-dependent information step by step. 
A projection layer and an Attention decoder are then followed to predict two streams of character sequences, corresponding to the two branches (CTC and Attention decoding) of the hybrid architecture.

Different from the baselines of CNVSRC 2023, we used a Bi-Transformer structure to implement the Attention decoder~\cite{wu2021u2++}, 
which consists of a 3-layer reverse Transformer and a 6-layer forward Transformer, settled upon the encoder in a parallel way. 
This modification enables the model to better learn the contextual relevance among characters within a sentence, 
thereby enhancing the accuracy of the prediction. 
During inference, the reverse Transformer is simply discarded, and only the forward Transformer is used to perform character prediction. 

The entire model was trained with a joint CTC/Attention loss~\cite{watanabe2017hybrid}, where the CTC loss is back-propagated through the projection layer, and the Attention loss is back-propagated through the Bi-Transformer decoder. 
Following~\cite{watanabe2017hybrid}, in the inference phase, the Beam Search approach is employed, which exclusively leverages the forward Transformer decoder and CTC for prediction. The allocation of weights during this process assigns 0.7 to the forward Transformer decoder and 0.3 to CTC. Notably, this implementation of beam search adopts a one-pass decoding strategy.
Additionally, adopting the experiences from CNVSRC 2023, we used Chinese character units (4,468 units) as the modeling units instead of subword units (5,904 units).

\begin{figure}[htbp]
  \centering
  \includegraphics[width=\linewidth]{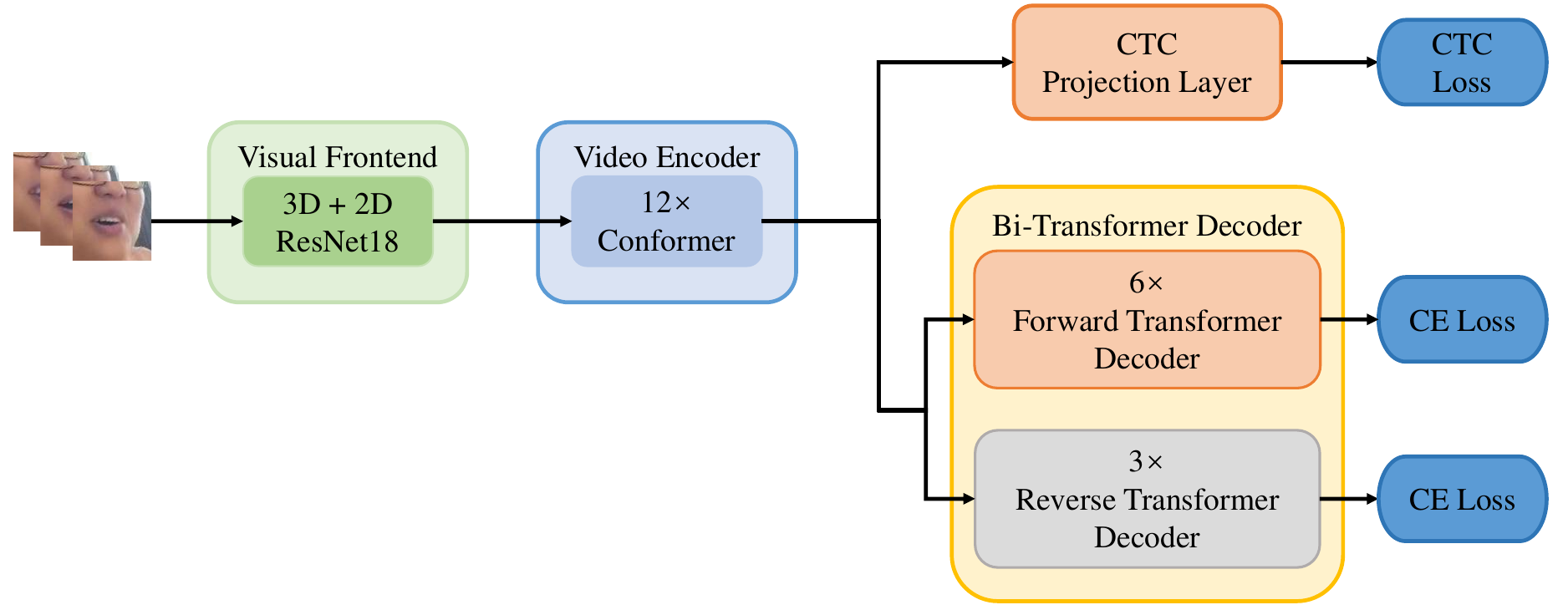}
  \caption{The model structure of CNVSRC 2024 baselines. Note that the reverse Transformer is discarded during inference; hence, it is drawn in grey.}
  \label{fig:model}
\end{figure}

\subsection{Data Preprocessing}

The videos of the provided datasets contain the entire faces of target speakers; thus, a preprocessing pipeline was employed to extract the lip region. 
This pipeline was applied to both the training data (CN-CVS) and the development/evaluation data (CNVSRC-Single and CNVSRC-Multi).
Since this pipeline was identical to that used in CNVSRC 2023, the readers can refer to~\cite{chen2024cn} for details.

\subsection{Training Strategy}

We followed the same two-step process as CNVSRC 2023 to build the baseline systems. 
Firstly, we performed pre-training with the CN-CVS dataset. 
Subsequently, the development set was split into an `adaptation set' and a `validation set', 
with a ratio of 8:1 for the single-speaker dataset and 3:1 for the multi-speaker dataset. 
The adaptation set was used to fine-tune the pre-trained model, while the validation set was used to select the appropriate checkpoint. 
The training process was summarized in Table~\ref{tab:training}, and the details are as follows. 

\begin{itemize}
\item \emph{P1}: The initial training phase (P1) selected the videos from CN-CVS with durations less than 4 seconds 
to train an `easy model'. 
The maximum number of training epochs was set to 75, and early stopping was triggered if the model's performance on the validation set began to decline. 
Once the P1 training stopped, we selected the top 10 models based on their accuracy on the validation set and averaged their parameters to obtain the P1 model. 

\item \emph{P2}: Next, a full pre-training phase (P2) was conducted using the complete CN-CVS dataset, with training conducted for 75 epochs. 
A warmup stage of 5 epochs was designed, during which the learning rate was gradually increased from 0 to 0.001.
The average of the models of the last 10 epochs was used as the P2 model, i.e., the pre-trained model.

\item \emph{FT}: The fine-tuning step started from the P2 model and ran 80 epochs, including two epochs of warmup. 
This process is the same for the models trained for the single-speaker and multi-speaker tasks, e.g., the average of the models of the last 5 epochs was used as the final model for both tasks. However, there are indeed some differences in the fine-tuning process. 
Firstly, the learning rate was slightly different, considering the different volumes of the training data (see Table~\ref{tab:training}). 
Moreover, considering the large amount of single-speaker data, we allowed more flexible model adaptation with the single-speaker task. 
To achieve this, the parameters of the classification layer of the single-speaker P2 model were re-randomized to provide sufficient space for fine-tuning. 
\end{itemize}

All these aforementioned models were trained using the AdamW optimizer~\cite{loshchilov2017decoupled}, 
with $\lambda=0.03, \beta_1=0.9, \beta_2=0.98$, representing the decay rate of the model parameters and their first and second moment, respectively.

\begin{table}[!htb]
\centering
\caption{Performance comparison of the baseline systems of CNVSRC 2024 and CNVSRC 2023.}
\vspace{-1mm}
\label{tab:result}
\resizebox{1\columnwidth}{!}{
\begin{tabular}{l|c|c|c|c}
    \toprule
    & \multicolumn{2}{c|}{T1: Single-speaker VSR} & \multicolumn{2}{c}{T2: Multi-speaker VSR} \\
    \cmidrule{2-5}
    & CNVSRC 2024 & CNVSRC 2023 & CNVSRC 2024 & CNVSRC 2023\\
    \midrule
    Valid~~ & 41.22\% & 48.57\% & 52.42\% & 58.77\% \\
    Eval & 39.66\% & 48.60\% & 52.20\% & 58.37\% \\
    \bottomrule
\end{tabular}}
\vspace{-2mm}
\end{table}

\subsection{Performance}
The performance of the baseline models was evaluated on the respective validation and evaluation sets for both the single-speaker task and multi-speaker task, 
using the TorchMetrics tool\footnote{https://lightning.ai/docs/torchmetrics/stable/}. 
The Character Error Rate (CER) results are shown in Table~\ref{tab:result}. 
Compared to the baseline of CNVSRC 2023, the baseline of CNVSRC 2024 demonstrates a notable performance improvement on both the single-speaker and multi-speaker tasks.

\section{CNVSRC 2024 Report}
\label{sec:res}

\begin{table*}[!htb]
\centering
\vspace{-5mm}
\caption{Leaderboard of CNVSRC 2024. TeamID and CER are reported.}
\vspace{-2mm}
\footnotesize
\label{tab:board}
\begin{tabular}{l|cc|cc|cc|cc}
\toprule
Task & \multicolumn{4}{c|}{T1: Single-speaker VSR} & \multicolumn{4}{c}{T2: Multi-speaker VSR} \\
\midrule
Track      & \multicolumn{2}{c|}{Fixed Track}&\multicolumn{2}{c|}{Open Track}&\multicolumn{2}{c|}{Fixed Track}&\multicolumn{2}{c}{Open Track} \\
\midrule
Baseline   &\multicolumn{2}{c|}{39.66\%}  &\multicolumn{2}{c|}{39.66\%}  &\multicolumn{2}{c|}{52.20\%}  &\multicolumn{2}{c}{52.20\%} \\
\midrule
Rank1 & T237 & 30.46\% & T170 & 30.06\% & T237 & 34.29\% & T237 & 34.29\% \\
Rank2 & T244 & 39.31\% & T237 & 30.46\% & T170 & 45.32\% & T170 & 38.34\% \\
Rank3 &      &         &      &         & T244 & 47.92\% & T405 & 57.77\% \\
\bottomrule
\end{tabular}
\vspace{-2mm}
\end{table*}

\subsection{Leaderboard}
CNVSRC 2024 received 10 valid submissions from 4 participating teams. 
Notably, in contrast to CNVSRC 2023, more teams this year focused on the multi-speaker track, indicating that VSR research is gradually shifting from single-speaker scenarios to more challenging multi-speaker scenarios. 
This shift highlights the growing importance of robust models that generalize well across speaker variations, a critical factor in real-world VSR applications.
The leaderboard results are reported in Table~\ref{tab:board}.

Among all participants, Team T237 achieved the best performance in 3/4 of tasks and tracks, while Team T170 demonstrated the strongest results in the T1 Open Track. 
All these teams significantly outperformed the baseline systems, achieving notable performance gains. 
Furthermore, a comparative analysis reveals that the relative improvement in the multi-speaker scenario is substantially larger than that in the single-speaker scenario, 
suggesting that recent advancements in LVS-VSR techniques are particularly effective in handling speaker variability.

\subsection{Technical Summary}

\subsubsection{Data Preprocessing and Augmentation}
Each team extended the baseline pipeline with advanced data preprocessing and augmentation strategies, aiming to improve model robustness against noise, pose variations, and other real-world challenges.
Team T237 followed the preprocessing methodology of their own CNVSRC 2023 system~\cite{wang2024npu}, which integrates random rotation, horizontal flipping, colour transformation, and multiple speed perturbations (e.g., 0.9× and 1.1×). 
These augmentation methods enhanced training data diversity, allowing the model to better generalize to variations in speaking speed and video quality. 
Additionally, they expanded the maximum crop size from 112 to 128 compared to their CNVSRC 2023 submission. This provides a larger view, enabling the model to capture subtle lip movements more accurately.

Team T244 introduced temporal masking as a core component of their preprocessing step. 
Instead of solely focusing on spatial augmentations, they applied masking along the temporal dimension. This forces the model to capture long-range temporal dependencies, enabling robust feature extraction even with missing frames caused by occlusions and motion blur.

\subsubsection{Frontend and Feature Extraction}
Some teams focused on advanced visual feature extraction.  
For instance, Team T237 improved the ResNet3D-based front-end~\cite{wang2024enhancing} by introducing a multi-scale feature extraction strategy. They incorporated an improved downsampling scheme alongside a larger input cropping size. 
This configuration enhanced the model's ability to extract fine-grained lip movement features, providing higher-quality visual representations for the subsequent encoder.  

Team T244 adopted a dual-path feature extraction strategy. They employed (1) ResNet3D to capture spatio-temporal dynamics from visual inputs and (2) ResNet1D to extract complementary audio features.
A Conformer module was subsequently employed to fuse these two modalities, fully leveraging the complementary nature of visual and audio cues to enhance speech recognition robustness.  

Team T405 used AV-HuBERT~\cite{shilearning} as a pre-trained feature extractor to obtain stable visual feature representations. In particular, they fused multi-layer outputs from AV-HuBERT into a Conformer-based architecture inspired by MFA-Conformer~\cite{zhang2022mfa}. This method demonstrated the usefulness of audio-visual pre-trained models. 
Although its overall performance was slightly lower than other approaches, it provides an interesting area for future research.

\subsubsection{Model Design}

Some teams applied advanced network structures and optimization techniques to improve performance.
For example, T237 and T405 adopted the E-Branchformer encoder~\cite{kim2023branchformer}.  
This parallel-branch architecture enables the model to simultaneously capture local and global dependencies, enhancing its capability to capture subtle variations in long-duration sequences.
Team T244 deviated from the standard Transformer-based decoder by introducing a memory-efficient S4D decoder~\cite{gu2022parameterization}. 
Unlike standard Transformer architectures, S4D excels in modeling long-term dependencies while maintaining computational efficiency~\cite{gu2021efficiently}.  
This architecture has already demonstrated strong performance in speech recognition and synthesis tasks~\cite{miyazaki2023structured}. 
The integration of S4D into the decoding module significantly enhanced the model’s temporal modeling capacity, improving its effectiveness in continuous visual speech recognition.

\subsubsection{Training Strategies and System Fusion}
To further improve generalization and model robustness, teams explored alternative training strategies and system fusion techniques.
Team T237 employed the Recognizer Output Voting Error Reduction (ROVER)~\cite{fiscus1997post} for post-fusion of multiple model outputs.  
Their ablation experiments demonstrated that incorporating ROVER led to significant CER reduction, highlighting the effectiveness of the ensemble method in VSR tasks.
Team T244 introduced a Kullback-Leibler (KL) divergence loss.  
It enforced consistency between the outputs of the forward and backward paths of their S4D decoder, which was supposed to improve the stability of the training process.

\section{Conclusion}
\label{sec:conc}
This paper presents a comprehensive overview of the second Chinese Continuous Visual Speech Recognition Challenge (CNVSRC 2024), highlighting its contributions to advancing Chinese LVC-VSR.  
Compared to CNVSRC 2023, this year’s challenge introduced a stronger baseline system and more training data, further pushing the boundary of the Chinese VSR research.
Based on the technical reports from the participants, we have summarized the key techniques that are crucial for constructing a strong continuous VSR system, 
including diverse data augmentation strategies, audio-visual pre-trained models for feature extraction, audio-visual feature fusion, 
memory-efficient S4D decoders, and ROVER-based system ensembles.  
Despite these advancements, the CER remains above 30\%, indicating that Chinese LVC-VSR still faces fundamental challenges 
before achieving real-world applicability.  

With the technical insights gained from this challenge, one can identify some important research directions. For example, appropriate
audio-visual alignment, self-supervised pre-training, and temporal augmentation. In particular, large language models (LLMs) may play an important role. A key issue of VSR is the uncertainty caused by weak information associated with lip movement, so linguistic and semantic constraints are crucial to obtaining reasonable performance. We expect that the massive amount of knowledge in LLMs will largely solve the problem and offer substantial performance improvement.

\newpage
\bibliographystyle{IEEEtran}
\bibliography{mybib}

\begin{thebibliography}{10}
\providecommand{\url}[1]{#1}
\csname url@samestyle\endcsname
\providecommand{\newblock}{\relax}
\providecommand{\bibinfo}[2]{#2}
\providecommand{\BIBentrySTDinterwordspacing}{\spaceskip=0pt\relax}
\providecommand{\BIBentryALTinterwordstretchfactor}{4}
\providecommand{\BIBentryALTinterwordspacing}{\spaceskip=\fontdimen2\font plus
\BIBentryALTinterwordstretchfactor\fontdimen3\font minus \fontdimen4\font\relax}
\providecommand{\BIBforeignlanguage}[2]{{%
\expandafter\ifx\csname l@#1\endcsname\relax
\typeout{** WARNING: IEEEtran.bst: No hyphenation pattern has been}%
\typeout{** loaded for the language `#1'. Using the pattern for}%
\typeout{** the default language instead.}%
\else
\language=\csname l@#1\endcsname
\fi
#2}}
\providecommand{\BIBdecl}{\relax}
\BIBdecl

\bibitem{goldschen1997continuous}
A.~J. Goldschen, O.~N. Garcia, and E.~D. Petajan, ``Continuous automatic speech recognition by lipreading,'' in \emph{Motion-Based recognition}.\hskip 1em plus 0.5em minus 0.4em\relax Springer, 1997, pp. 321--343.

\bibitem{tye2007audiovisual}
N.~Tye-Murray, M.~S. Sommers, and B.~Spehar, ``Audiovisual integration and lipreading abilities of older adults with normal and impaired hearing,'' \emph{Ear and hearing}, vol.~28, no.~5, pp. 656--668, 2007.

\bibitem{hassanat2011visual}
A.~B. Hassanat, ``Visual speech recognition,'' \emph{Speech and Language Technologies}, vol.~1, pp. 279--303, 2011.

\bibitem{haliassos2021lips}
A.~Haliassos, K.~Vougioukas, S.~Petridis, and M.~Pantic, ``Lips don't lie: A generalisable and robust approach to face forgery detection,'' in \emph{Proceedings of the IEEE/CVF conference on computer vision and pattern recognition}, 2021, pp. 5039--5049.

\bibitem{yeo2023multi}
J.~H. Yeo, M.~Kim, and Y.~M. Ro, ``Multi-temporal lip-audio memory for visual speech recognition,'' in \emph{ICASSP 2023-2023 IEEE International Conference on Acoustics, Speech and Signal Processing (ICASSP)}.\hskip 1em plus 0.5em minus 0.4em\relax IEEE, 2023, pp. 1--5.

\bibitem{ma2022training}
P.~Ma, Y.~Wang, S.~Petridis, J.~Shen, and M.~Pantic, ``Training strategies for improved lip-reading,'' in \emph{ICASSP 2022-2022 IEEE International Conference on Acoustics, Speech and Signal Processing (ICASSP)}.\hskip 1em plus 0.5em minus 0.4em\relax IEEE, 2022, pp. 8472--8476.

\bibitem{yang2022improved}
H.~Yang, T.~Luo, Y.~Zhang, M.~Song, L.~Xie, Y.~Yan, and E.~Yin, ``Improved word-level lipreading with temporal shrinkage network and netvlad,'' in \emph{Proceedings of the 2022 International Conference on Multimodal Interaction}, 2022, pp. 504--508.

\bibitem{chung2017lip}
J.~S. Chung and A.~Zisserman, ``Lip reading in the wild,'' in \emph{Computer Vision--ACCV 2016: 13th Asian Conference on Computer Vision, Taipei, Taiwan, November 20-24, 2016, Revised Selected Papers, Part II 13}.\hskip 1em plus 0.5em minus 0.4em\relax Springer, 2017, pp. 87--103.

\bibitem{kim2022distinguishing}
M.~Kim, J.~H. Yeo, and Y.~M. Ro, ``Distinguishing homophenes using multi-head visual-audio memory for lip reading,'' in \emph{Proceedings of the AAAI conference on artificial intelligence}, vol.~36, no.~1, 2022, pp. 1174--1182.

\bibitem{chang2024conformer}
O.~Chang, H.~Liao, D.~Serdyuk, A.~Shahy, and O.~Siohan, ``Conformer is all you need for visual speech recognition,'' in \emph{ICASSP 2024-2024 IEEE International Conference on Acoustics, Speech and Signal Processing (ICASSP)}.\hskip 1em plus 0.5em minus 0.4em\relax IEEE, 2024, pp. 10\,136--10\,140.

\bibitem{serdyuk2022transformer}
D.~Serdyuk, O.~Braga, and O.~Siohan, ``Transformer-based video front-ends for audio-visual speech recognition for single and multi-person video,'' in \emph{INTERSPEECH}, 2022, pp. 2833--2837.

\bibitem{shilearning}
B.~Shi, W.-N. Hsu, K.~Lakhotia, and A.~Mohamed, ``Learning audio-visual speech representation by masked multimodal cluster prediction,'' in \emph{International Conference on Learning Representations}.

\bibitem{shillingford2018large}
B.~Shillingford, Y.~Assael, M.~W. Hoffman, T.~Paine, C.~Hughes, U.~Prabhu, H.~Liao, H.~Sak, K.~Rao, L.~Bennett \emph{et~al.}, ``Large-scale visual speech recognition,'' in \emph{INTERSPEECH}, 2019, pp. 4135--4139.

\bibitem{chen2023cn}
C.~Chen, D.~Wang, and T.~F. Zheng, ``{CN-CVS}: A {M}andarin audio-visual dataset for large vocabulary continuous visual to speech synthesis,'' in \emph{ICASSP 2023-2023 IEEE International Conference on Acoustics, Speech and Signal Processing (ICASSP)}.\hskip 1em plus 0.5em minus 0.4em\relax IEEE, 2023, pp. 1--5.

\bibitem{chen2024cn}
C.~Chen, Z.~Liu, X.~Li, L.~Li, and D.~Wang, ``{CNVSRC} 2023: The first {C}hinese continuous visual speech recognition challenge,'' in \emph{INTERSPEECH}, 2024, pp. 1930--1934.

\bibitem{ma2023auto}
P.~Ma, A.~Haliassos, A.~Fernandez-Lopez, H.~Chen, S.~Petridis, and M.~Pantic, ``{Auto-AVSR}: Audio-visual speech recognition with automatic labels,'' in \emph{ICASSP 2023-2023 IEEE International Conference on Acoustics, Speech and Signal Processing (ICASSP)}.\hskip 1em plus 0.5em minus 0.4em\relax IEEE, 2023, pp. 1--5.

\bibitem{stafylakis2017combining}
T.~Stafylakis and G.~Tzimiropoulos, ``Combining residual networks with {LSTMs} for lipreading,'' in \emph{INTERSPEECH}, 2017, pp. 3652--3656.

\bibitem{gulati2020conformer}
A.~Gulati, J.~Qin, C.-C. Chiu, N.~Parmar, Y.~Zhang, J.~Yu, W.~Han, S.~Wang, Z.~Zhang, Y.~Wu \emph{et~al.}, ``Conformer: Convolution-augmented {T}ransformer for speech recognition,'' in \emph{INTERSPEECH}, 2020, pp. 5036--5040.

\bibitem{wu2021u2++}
D.~Wu, B.~Zhang, C.~Yang, Z.~Peng, W.~Xia, X.~Chen, and X.~Lei, ``U2++: Unified two-pass bidirectional end-to-end model for speech recognition,'' \emph{arXiv preprint arXiv:2106.05642}, 2021.

\bibitem{watanabe2017hybrid}
S.~Watanabe, T.~Hori, S.~Kim, J.~R. Hershey, and T.~Hayashi, ``Hybrid {CTC}/attention architecture for end-to-end speech recognition,'' \emph{IEEE Journal of Selected Topics in Signal Processing}, vol.~11, no.~8, pp. 1240--1253, 2017.

\bibitem{loshchilov2017decoupled}
I.~Loshchilov and F.~Hutter, ``Decoupled weight decay regularization,'' in \emph{International Conference on Learning Representations}, 2019.

\bibitem{wang2024npu}
H.~Wang, P.~Guo, W.~Chen, P.~Zhou, and L.~Xie, ``The {NPU-ASLP-LiAuto} system description for visual speech recognition in {CNVSRC} 2023,'' \emph{arXiv preprint arXiv:2401.06788}, 2024.

\bibitem{wang2024enhancing}
H.~Wang, P.~Guo, X.~Wan, H.~Zhou, and L.~Xie, ``Enhancing lip reading with multi-scale video and multi-encoder,'' \emph{arXiv preprint arXiv:2404.05466}, 2024.

\bibitem{zhang2022mfa}
Y.~Zhang, Z.~Lv, H.~Wu, S.~Zhang, P.~Hu, Z.~Wu, H.-y. Lee, and H.~Meng, ``{MFA-Conformer}: Multi-scale feature aggregation conformer for automatic speaker verification,'' in \emph{INTERSPEECH}, 2022, pp. 306--310.

\bibitem{kim2023branchformer}
K.~Kim, F.~Wu, Y.~Peng, J.~Pan, P.~Sridhar, K.~J. Han, and S.~Watanabe, ``E-branchformer: Branchformer with enhanced merging for speech recognition,'' in \emph{2022 IEEE Spoken Language Technology Workshop (SLT)}.\hskip 1em plus 0.5em minus 0.4em\relax IEEE, 2023, pp. 84--91.

\bibitem{gu2022parameterization}
A.~Gu, K.~Goel, A.~Gupta, and C.~R{\'e}, ``On the parameterization and initialization of diagonal state space models,'' \emph{Advances in Neural Information Processing Systems}, vol.~35, pp. 35\,971--35\,983, 2022.

\bibitem{gu2021efficiently}
A.~Gu, K.~Goel, and C.~R{\'e}, ``Efficiently modeling long sequences with structured state spaces,'' in \emph{International Conference on Learning Representations}, 2022.

\bibitem{miyazaki2023structured}
K.~Miyazaki, M.~Murata, and T.~Koriyama, ``Structured state space decoder for speech recognition and synthesis,'' in \emph{ICASSP 2023-2023 IEEE International Conference on Acoustics, Speech and Signal Processing (ICASSP)}.\hskip 1em plus 0.5em minus 0.4em\relax IEEE, 2023, pp. 1--5.

\bibitem{fiscus1997post}
J.~G. Fiscus, ``A post-processing system to yield reduced word error rates: Recognizer output voting error reduction (rover),'' in \emph{1997 IEEE Workshop on Automatic Speech Recognition and Understanding Proceedings}.\hskip 1em plus 0.5em minus 0.4em\relax IEEE, 1997, pp. 347--354.

\end{thebibliography}

\end{document}